\documentclass[runningheads]{llncs}

\usepackage{eccv}

\usepackage{eccvabbrv}

\usepackage{graphicx}
\usepackage{booktabs}
\usepackage{wrapfig}
\usepackage{subcaption}
\usepackage{caption}
\usepackage{tabularx}

\usepackage[accsupp]{axessibility}  

\usepackage{hyperref}

\usepackage{orcidlink}
\usepackage{color}
\usepackage{xcolor}
\usepackage{multirow}
\usepackage{setspace}
\usepackage{array} 
\usepackage{amsmath}
\usepackage[ruled,vlined,linesnumbered,noend]{algorithm2e}

\SetAlFnt{\small}
\SetAlCapFnt{\normalsize}
\SetAlCapNameFnt{\normalsize}
\SetKwInput{KwIn}{Require}
\SetKwInput{KwOut}{Ensure}
\DontPrintSemicolon

\begin{document}

\title{Self-Evolving Agentic Image Restoration via Deliberate Planning and Intuitive Execution} 

\titlerunning{Self-Evolving Agentic Image Restoration}

\author{Shuang Cui\inst{1,2}\orcidlink{0000-0001-9293-7316} \and
Fan Ji\inst{1}\orcidlink{0009-0007-3068-4158} \and
Guanglong Sun\inst{3}\orcidlink{0009-0001-8403-1925} \and Yufei Guo\inst{4} \and \\ Xiongxin Tang\inst{1}$^{\dagger}$ \and Jiangmeng Li\inst{1}$^{\dagger}$\orcidlink{0000-0002-3376-1522} \and Fanjiang Xu\inst{1}\orcidlink{0009-0004-6016-7360} }

\authorrunning{S.~Cui et al.}

\institute{Institute of Software, Chinese Academy of Sciences, Beijing, China
\and
University of Chinese Academy of Sciences, Beijing, China
\and
School of Life Sciences, Tsinghua University, Beijing, China\\
\and
Intelligent Science \& Technology Academy of CASIC, Beijing, China \\
\email{\{cuishuang21,jifan22\}@mails.ucas.ac.cn},
\email{sgl23@mails.tsinghua.edu.cn},
\email{yfguo@pku.edu.cn},
\email{\{xiongxin,jiangmeng2019,fanjiang\}@iscas.ac.cn}\\
$^{\dagger}$Corresponding authors
}

\maketitle

\begin{abstract}
  Real-world image restoration (IR) remains challenging due to complex and coupled degradations. While recent agentic IR frameworks leverage Large Language Models for flexible tool planning, they face two critical limitations. First, from a search scheme perspective, excessive reliance on greedy strategies fails to balance exploration and exploitation. Second, existing agentic systems underutilize information, exhibiting episodic amnesia. To address these challenges, we propose \textbf{Self-Evolving Agentic Image Restoration (SEAR)}, which formulates restoration as a sequential decision-making problem. Inspired by the dual-process theory, SEAR comprises an Intuitive Executor and a Deliberate Planner, respectively following the fast-thinking \textit{System 1} and slow-thinking \textit{System 2} principles. The Deliberate Planner employs Pruning-Aware Monte Carlo Tree Search for long-horizon reasoning, utilizing a hybrid no-reference reward and a Multimodal Large Language Model (MLLM)-based tournament to prevent metric exploitation. Complementarily, the Intuitive Executor leverages a self-evolving episodic memory indexed by degradation-aware state fingerprints. This mechanism distills expensive search trajectories into adaptive expertise, overcoming episodic amnesia while progressively amortizing cold-start exploration costs through memory reuse. Extensive experiments on synthetic and real-world benchmarks demonstrate its strong perceptual and quantitative performance.
\end{abstract}

\section{Introduction}
\label{sec:intro}

\begin{figure}[tb]
  \centering
\includegraphics[width=\textwidth]{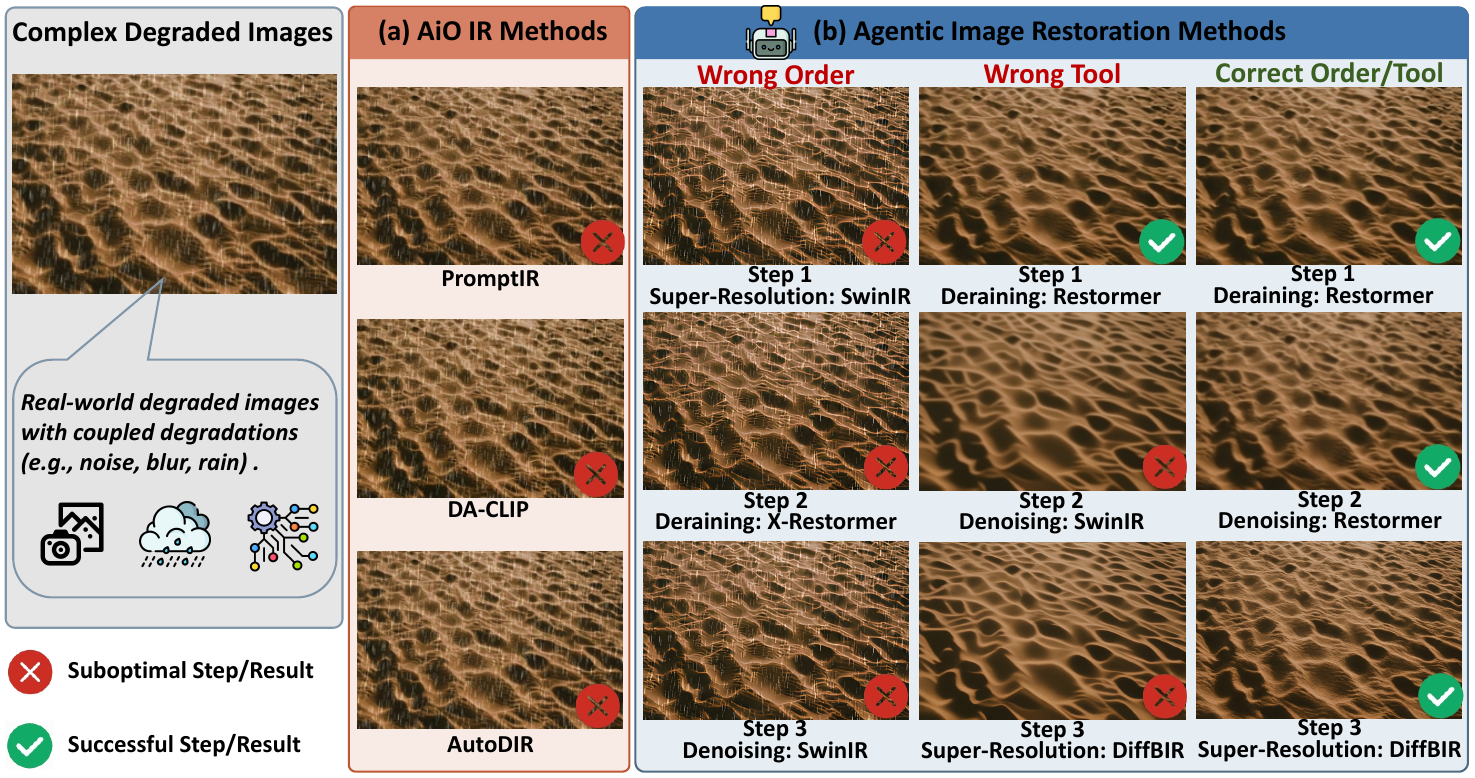}
  \caption{\textbf{Analysis of restoration limitations.} (a) Static AiO models struggle with coupled degradations. (b) Restoration quality is highly sensitive to execution sequences and tools, where greedy decisions (e.g., over-smoothed) incur irreversible failure.}
  \label{FIG_limitation}
\end{figure}

Image Restoration (IR) is a fundamental task in computer vision that focuses on recovering high-quality images from degraded observations \cite{zhang2017beyond,pei2018does,zamir2022restormer,zou2023object,jiang2025survey}. Task-specific networks have achieved remarkable success in handling isolated degradations such as denoising \cite{chen2023masked,tian2023multi}, deblurring \cite{zamir2022restormer,heblurdm2025,cui2025causiam}, and super-resolution \cite{lin2024diffbir,wu2024one,pmlr-v267-qu25h}. However, in real-world scenarios, images exhibit complex and coupled degradations \cite{wei2020physics, jiang2025multi}. All-in-One (AiO) IR frameworks \cite{li2022all,potlapalli2023promptir,jiang2024autodir,wang2026all} attempt to handle multiple tasks within a unified model, yet these static designs struggle to generalize to unknown distortion combinations \cite{jiang2025survey}, as shown in Fig.~\ref{FIG_limitation} (a).

Recently, inspired by the advanced reasoning and planning capabilities of Large Language Models (LLM) \cite{wei2022chain,touvron2023llama,hui2024qwen2}, agentic restoration frameworks \cite{chen2024restoreagent,zhuintelligent2025,zuo4kagent2025} have emerged as a powerful paradigm for complex image restoration. As a representative example, AgenticIR \cite{zhuintelligent2025} distills statistical experience from tool sequence traversals into LLM prompts to guide the agent, thereby improving restoration consistency. Despite these advances, existing agentic approaches suffer from structural limitations that hinder performance in challenging scenarios.

\begin{figure}[tb]
  \centering
  \includegraphics[width=\textwidth]{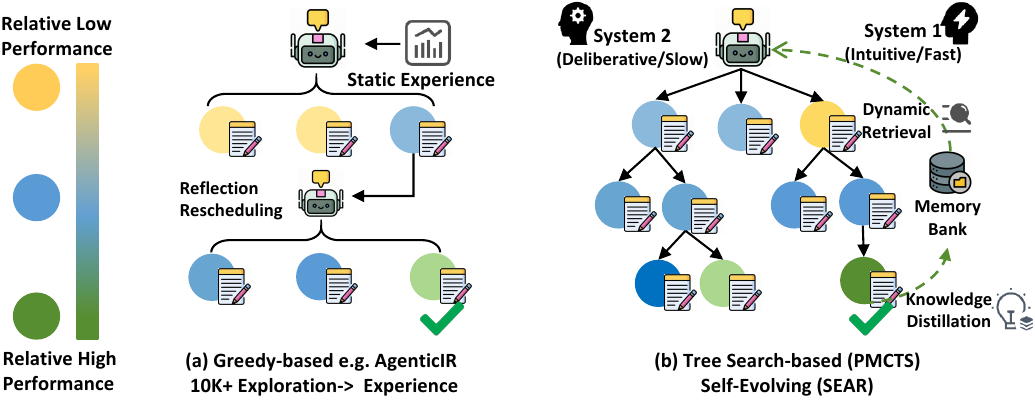}
  \caption{\textbf{Comparison of restoration decision strategies.} (a) Myopic greedy agents prioritize local gains, yielding suboptimal trajectories. (b) SEAR synergizes an Intuitive Executor with a Deliberate Planner for long-horizon trajectory optimization.}
  \label{FIG_comparison}
\end{figure}

From a search scheme perspective, current agents \cite{zhuintelligent2025,zuo4kagent2025} predominantly rely on greedy heuristics that lack long-horizon foresight and fail to balance exploration and exploitation in high-dimensional decision spaces. By prioritizing immediate rewards at each step (Fig.~\ref{FIG_comparison} (a)), these methods often converge to sub-optimal restoration trajectories \cite{yuexact2025,gaorf}. As shown in Fig.~\ref{FIG_limitation} (b), aggressively applying denoising to maximize intermediate cleanliness may irreversibly destroy fine structural details that are essential for subsequent super-resolution, resulting in over-smoothed outputs. Moreover, from the perspective of information utility, existing systems exhibit episodic amnesia. Current agents rely primarily on static prompt-level experience, while high-reward trajectories discovered through computationally expensive search are treated as transient and discarded upon completion of each image restoration. Consequently, knowledge acquired from previous exploration cannot be transferred across diverse cases \cite{park2023generative,zhong2024memorybank}, forcing the agent to repeatedly resolve similar degradation patterns from scratch and increasing repeated search costs in practical deployments.   

To address these challenges, we reformulate real-world image restoration as a long-horizon sequential decision-making problem, where restoration operators must be applied in a carefully coordinated order to preserve image structures. This perspective moves beyond greedy, step-wise decisions to global path-level reasoning. Inspired by dual-process theory \cite{stanovich1999rational,kahneman2011thinking}, we propose \textbf{Self-Evolving Agentic Image Restoration (SEAR)}. SEAR employs an Intuitive Executor and a Deliberate Planner to emulate the cognitive synergy between fast \textit{System 1} execution and slow \textit{System 2} reasoning. As shown in Fig.~\ref{FIG_comparison}, SEAR replaces traditional step-wise heuristics with a self-evolving architecture that continuously refines restoration trajectories by dynamically accumulating experience.

Specifically, the Deliberate Planner performs hierarchical reasoning, leveraging LLM for macro-level task scheduling and Pruning-Aware Monte Carlo Tree Search (P-MCTS) for micro-level tool coordination, balancing exploration and exploitation. P-MCTS trajectories are evaluated via a hybrid no-reference reward and safeguarded by a Multimodal Large Language Model (MLLM)-based perceptual tournament to prevent adversarial metric exploitation and ensure human-aligned fidelity. Complementarily, the Intuitive Executor addresses insufficient information utilization, i.e., episodic amnesia in existing agentic restoration systems. It maintains a self-evolving episodic memory that distills high-reward trajectories into reusable execution strategies indexed by state fingerprints. When structurally similar degradation patterns recur, the Intuitive Executor retrieves and executes these trajectories instead of replanning from scratch, effectively reusing past exploration results. Through continuous self-evolution, SEAR transforms transient search experiences into adaptive knowledge, amortizes repeated planning costs for recurring degradation patterns, and gradually shifts restoration from search-driven reasoning to memory-guided execution.
Main contributions are summarized as follows:
\begin{itemize}
\item We propose SEAR, a dual-process restoration agent that synergizes deliberate planning with intuitive execution to tackle complex image restoration.
\item We introduce P-MCTS, leveraging MLLM-guided long-horizon rollouts to coordinate tool selection while balancing exploration and exploitation.
\item We develop a self-evolving episodic memory that converts search experiences into retrievable, adaptive knowledge, resolving amnesia and reducing exploration overhead. 
\item Extensive experiments on synthetic and real-world datasets demonstrate that SEAR achieves state-of-the-art perceptual quality, validating a robust paradigm evolving from costly deliberation to memory-guided execution.
\end{itemize}


\section{Related work}
\label{sec:related_work}

\paragraph{\textbf{Image Restoration.}} Image Restoration (IR) transitions from task-specific architectures for denoising \cite{chen2023masked,tian2023multi}, deblurring \cite{zamir2022restormer,heblurdm2025,cui2025causiam}, deraining \cite{chen2023learning,xiao2022image}, super-resolution \cite{lin2024diffbir,wu2024one,pmlr-v267-qu25h} toward All-in-One (AiO) IR frameworks that handle multiple degradations within a unified model. Representative methods include AirNet \cite{li2022all}, which employs contrastive learning for feature discrimination, as well as PromptIR \cite{potlapalli2023promptir} and InstructIR \cite{conde2024instructir}, which utilize degradation-aware contexts or human instructions to guide restoration. Advancing this paradigm, AutoDIR \cite{jiang2024autodir} integrates semantic-agnostic VLMs to automate degradation identification and steer structure-aware diffusion processes. Furthermore, DA-CLIP \cite{luo2024controlling} and MiOIR \cite{kong2024towards} leverage pre-trained foundation model priors or sequential prompting to enhance robustness. Despite these advancements, AiO methods struggle to balance generalization and reconstruction. Their efficacy remains limited on out-of-distribution degradation combinations \cite{jiang2025survey,duan2025test} due to a lack of strategic reasoning for intricate real-world scenarios.

\paragraph{\textbf{Autonomous Agents in Image Restoration.}} Large Language Models (LLM) \cite{wei2022chain,touvron2023llama,hui2024qwen2} exhibit formidable logical reasoning and task-planning capabilities, driving a paradigm shift from static restoration models toward adaptive, agentic systems \cite{durante2024agent,hong2023metagpt,wu2024autogen}. Foundational works such as RestoreAgent \cite{chen2024restoreagent} and AgenticIR \cite{zhuintelligent2025} utilize in-context learning for tool dispatching and visual-linguistic feedback for iterative calibration. Multi-agent frameworks such as 4KAgent \cite{zuo4kagent2025}, MAIR \cite{jiang2025multi}, and HybridAgent \cite{li2025hybrid} address coupled degradations through specialized sub-agents or expert integration. While training-intensive methods like JarvisIR \cite{lin2025jarvisir} enhance MLLM via supervision or alignment, they require substantial training overhead. In contrast, SEAR follows a reasoning-centric paradigm without additional training. 

\noindent Despite their adaptability, existing agentic frameworks \cite{zhuintelligent2025,zuo4kagent2025,jiang2025multi,li2025hybrid} are often constrained by myopic execution strategies that fail to balance exploration and exploitation. Furthermore, these methods suffer from diminished information utility, as they treat successful reasoning trajectories as ephemeral context rather than persistent assets. In the absence of long-horizon foresight and structural memory, such agents are trapped in a non-evolving loop, redundantly resolving recurring degradation patterns from scratch. SEAR addresses these gaps by combining hierarchical planning with a self-evolving episodic memory, systematically converting successful trajectories into reusable, long-term expertise.    

\section{Methods}
\label{sec:methods}

\begin{figure}[t!]
  \centering
  \includegraphics[width=\textwidth]{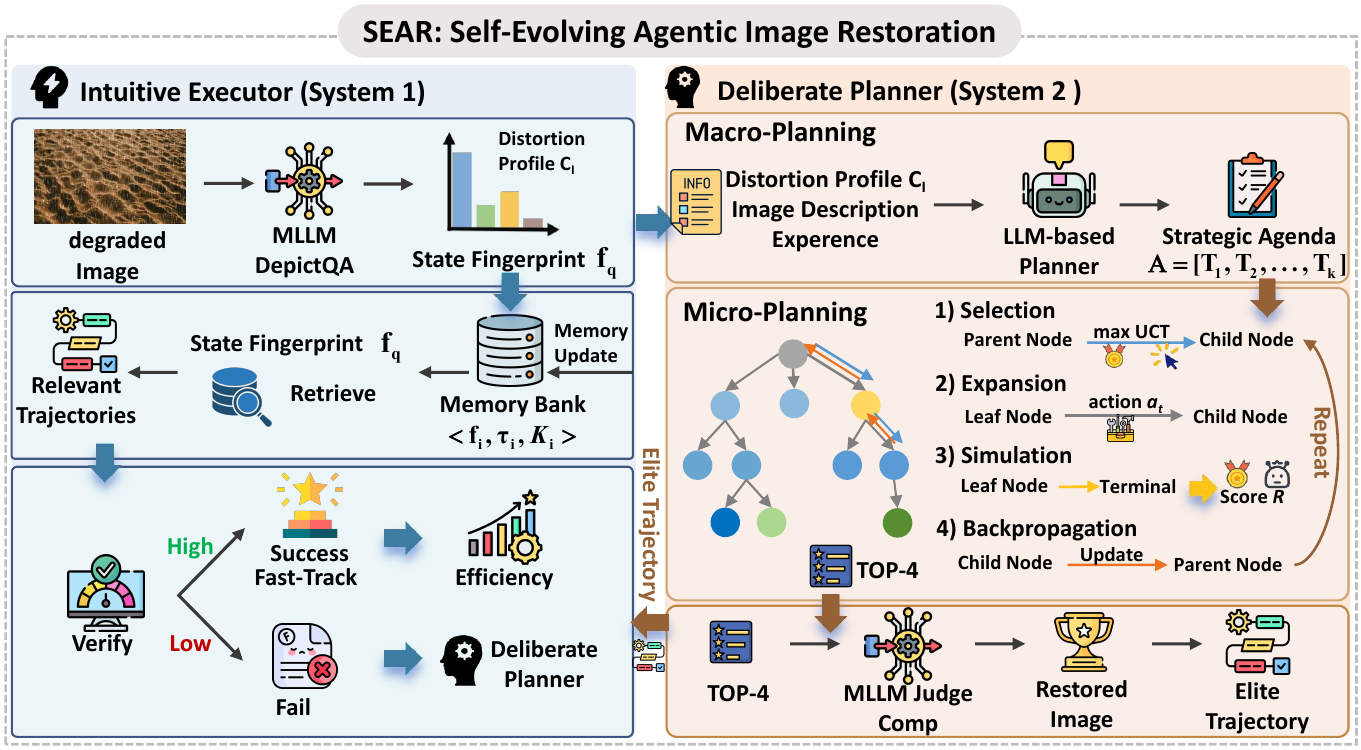}
  \caption{\textbf{Overview of the SEAR Framework.} SEAR follows a dual-process design: the Intuitive Executor performs rapid, memory-driven inference by retrieving proven strategies, while the Deliberate Planner tackles complex scenarios via long-horizon hierarchical optimization. By progressively distilling planned trajectories into the Executor, SEAR evolves from costly deliberation to memory-guided execution.}
  \label{FIG_framework}
\end{figure}

We propose SEAR, a dual-process architecture for agentic image restoration (Fig.~\ref{FIG_framework}). To overcome the limitations of myopic and memory-less agents, SEAR is decoupled into an Intuitive Executor (\textit{System 1}) for rapid, memory-driven inference, and a Deliberate Planner (\textit{System 2}) for slow, long-horizon exploration. A self-evolving memory bank bridges the two modules by consolidating high-reward trajectories into persistent expertise, as detailed in Appendix D.

\subsection{Problem Formulation}\label{sec:problem_formulation}

\paragraph{\textbf{State and Action Space.}} Real-world image restoration is inherently order-sensitive (e.g., denoising before deblurring leads to different structural outcomes). Let $\mathcal{A} = \{f_1, \dots, f_K\}$ denote the restoration tool library and $I_0$ the initial degraded image. A restoration trajectory is defined as an ordered sequence $\tau = (a_0, \dots, a_{H-1})$, where each $a_t \in \mathcal{A}$, $H$ denotes the trajectory horizon (i.e., the number of sequential operations), and $I_{t+1} = a_t(I_t)$ represents the state transition after applying $a_t$. By explicitly modeling the full trajectory, this formulation accounts for delayed effects across steps, ensuring that early decisions do not irreversibly compromise structures required by subsequent operations.

\paragraph{\textbf{No-Referenced Reward Objective.}} Since paired ground truths are unavailable in real-world inference, evaluating long-horizon trajectories relies solely on blind quality assessment. To prevent metric overfitting that can produce misleading artifacts, we guide the search with a robust perceptual ensemble. The optimal trajectory $\tau^*$ maximizes a terminal hybrid reward $R(I_H)$, defined as a weighted combination of human-aligned aesthetics \cite{wu2023human} and no-reference IQA metrics:
\begin{equation}
\tau^* = \arg\max_\tau R(I_H), \quad 
R(I_H) = \lambda_h \, \mathrm{HPSv2}(I_H) + \sum_{j\in\mathcal{J}} \lambda_j \, \widehat{\mathrm{IQA}}_j(I_H),
\label{equ_reward}
\end{equation}
where $I_H$ denotes the terminal restored image of trajectory $\tau$. The IQA ensemble includes NIQE \cite{mittal2012making}, MANIQA \cite{yang2022maniqa}, MUSIQ \cite{ke2021musiq}, and CLIP-IQA \cite{wang2023exploring}. Each IQA metric $\widehat{\mathrm{IQA}}_j$ is normalized to zero mean and unit variance. The specific weight configurations for the reward ensemble follow the settings established in \cite{zuo4kagent2025}. We denote $R(\tau)\equiv R(I_H(\tau))$ for the reward of trajectory $\tau$. This approach ensures perceptual consistency over single-metric optimization (Sec.~\ref{subsection:ablation_study}).

\subsection{Intuitive Executor with Self-Evolving Memory} 
\label{sec:intuitive_executor}
Unlike prior agents that suffer episodic amnesia by discarding trajectories post-inference, the Intuitive Executor leverages a self-evolving memory bank $\mathcal{M}$. To support adaptive execution, $\mathcal{M}$ is formalized as a repository of expertise tuples $\langle \mathbf{f}_i, \tau_i, \mathcal{K}_i \rangle$, where each state-trajectory pair is linked to a statistical profile $\mathcal{K}_i = \{n, \mu, R_{\max}\}$ tracking its visit count, mean reward, historical peak reward.

\paragraph{\textbf{Degradation-Aware State Fingerprint.}} 
To facilitate efficient and content-invariant memory addressing, we first project the high-dimensional degraded image $I_0$ into a structured state fingerprint $\mathbf{f}$. A fine-tuned MLLM \cite{you2024depicting,zhuintelligent2025} diagnoses the input $I_0$ to extract a semantic distortion profile $C_I$, which is quantized into a state vector $\mathbf{f} = [l_1, \dots, l_{|\mathcal{D}|}]^\top$. Here, $|\mathcal{D}|=8$ spans a predefined taxonomy of degradation primitives (e.g., blur, noise, compression artifacts), with each element $l_i \in {0, \dots, 5}$ encoding discrete intensity levels. This coarse-grained quantization acts as a regularization mechanism. By preventing the memory bank from overfitting to negligible pixel-level variances, it ensures that structurally identical degradation patterns are robustly mapped to the same execution strategy, thereby providing a low-dimensional key for efficient retrieval.

\paragraph{\textbf{Potential-Driven Retrieval.}} 
Given the query fingerprint $\mathbf{f}_q$, the Intuitive Executor retrieves $\mathcal{M}$ to identify a candidate set $\mathcal{S}_{\mathrm{cand}}$ composed of entries within an L2 proximity threshold. To balance prior performance with state similarity, the agent ranks these candidates using a potential-aware score:
\begin{equation}
\mathcal{C}(\tau_i) = R_{\max}(\tau_i) - \alpha \|\mathbf{f}_q - \mathbf{f}_i\|_2,
\end{equation}
where $R_{\max}(\tau_i)$ is the historical peak reward of $\tau_i$, and $\alpha$ penalizes state discrepancy. By focusing on $R_{\max}$ rather than the mean reward $\mu$, the agent prioritizes trajectories with the highest demonstrated restorative potential, thereby reducing sensitivity to noisy IQA estimates and suboptimal historical averages.

\paragraph{\textbf{Reliability-Gated Adaptive Execution.}}
To preserve restoration fidelity, we introduce an outcome-driven verification gate for retrieved candidate trajectories. The top-ranked retrieved trajectory $\tau_{\mathrm{ret}}$ is first executed to generate a candidate restoration $I_H(\tau_{\mathrm{ret}})$, which is then evaluated by the hybrid reward function in Eq.~\ref{equ_reward}. If the reward meets the shortcut threshold $\zeta$, the agent accepts this trajectory and outputs $I_H(\tau_{\mathrm{ret}})$ as the final result. Otherwise, the retrieved trajectory is used only as a warm start for the Deliberate Planner, which further performs P-MCTS search. This gate allows shortcut execution only for reliable retrieved trajectories, reducing unnecessary planning while preserving a quality-oriented fallback mechanism.

\subsection{Deliberate Planner for Long-Horizon Reasoning}
\label{subsec:deliberate_planner}
The Deliberate Planner is invoked when the Intuitive Executor lacks reliable precedents. To handle the combinatorial action space, it employs a hierarchical planning strategy, separating macro-level agenda reasoning from micro-level tool instantiation.

\paragraph{\textbf{Macro-Planning: Agenda-Driven Space Pruning.}}
Guided by the distortion profile $C_I$ and prior restoration experience \cite{zhuintelligent2025}, an LLM generates a strategic agenda $\mathcal{T}=[T_1,\dots,T_L]$. This ordered task sequence encodes intrinsic restoration priorities (e.g., denoising before super-resolution) and enforces precedence constraints, reducing the exponential search space of joint tool selection into a stage-wise optimization sequence for subsequent tree search.

\paragraph{\textbf{Micro-Planning: Pruning-Aware Monte Carlo Tree Search.}} 
To instantiate the strategic agenda into a concrete tool sequence, the Deliberate Planner employs Pruning-Aware Monte Carlo Tree Search (P-MCTS). This framework shifts the optimization target from isolated tool decisions to long-horizon trajectory optimization. In the search tree, each node $s$ represents an intermediate restoration state, while each edge signifies the invocation of an expert tool $a \in \text{Tools}(T_i)$. P-MCTS identifies high-reward trajectories through an iterative process of Selection, Expansion, Simulation, and Backpropagation. The search runs for $N_{total} = \omega \cdot L$ iterations, where $L$ is the agenda length and $\omega$ is the search budget multiplier. In practice, this process is significantly accelerated by a caching mechanism that enables the rapid reuse of previously explored results. This cycle, detailed below, balances exploration and exploitation to converge toward globally optimal restoration outcomes.

\noindent \textbf{Selection}: Starting from the root, the agent traverses the tree by selecting the child node $s_{child}$ that maximizes the Upper Confidence Bound for Trees (UCT):
\begin{equation}
UCT(s_{child}) = \frac{Q(s_{child})}{N(s_{child})} + c \cdot \sqrt{\frac{2 \ln N(s_{parent})}{N(s_{child})}},
\end{equation}
where $Q$ represents the accumulated reward, $N$ denotes the visit count, and $c$ is the exploration parameter. This strategy prioritizes successful paths while ensuring sufficient coverage of the decision space.

\noindent \textbf{Expansion}: Upon reaching a node $s_n$ that is not yet fully expanded for the current subtask $T_i$, the agent expands the tree by sampling an operator $a \in \text{Tools}(T_i)$ to generate a child node $s_{\text{child}}$. This agenda-constrained expansion restricts the branching factor to operators associated with the current subtask, enforcing the strategic topology defined by the agenda. As a result, the search space is substantially pruned compared to unconstrained tool selection.

\noindent \textbf{Simulation}: To estimate the downstream impact of current decisions, the agent performs a look-ahead rollout by sequentially applying tools from the remaining subtasks $[T_{i+1}, \dots, T_L]$ to reach a terminal state $s_T$. The resulting terminal image $I_H$ is then evaluated via the hybrid reward $R(I_H)$ as defined in Eq.~\ref{equ_reward}.

\noindent \textbf{Backpropagation}: The terminal reward $R(I_H)$ is backpropagated to update the statistics of all ancestral nodes $s_j$ along the trajectory:
\begin{equation}
N(s_j) \leftarrow N(s_j) + 1, Q(s_j) \leftarrow Q(s_j) + R(I_H), V_{\max}(s_j) \leftarrow \max(V_{\max}(s_j), R(I_H)).
\end{equation}
This iterative refinement steers the selection strategy toward the high-reward sequence of expert tools. By tracking the maximum value $V_{\max}$, the framework preserves the highest potential of explored paths to guide future search iterations.

\paragraph{\textbf{Perceptual Alignment via MLLM Tournament.}}
To mitigate the risk of metric exploitation, where restoration operators may generate adversarial artifacts yielding deceptively high scores, we implement a pairwise tournament verification. The top four candidate trajectories $\{\tau^{(1)},\dots,\tau^{(4)}\}$, ranked according to their terminal rewards $R(\tau^{(i)})$, undergo pairwise visual assessments by an MLLM judge. Acting as a perceptual safety net, this final check goes beyond the hybrid reward. Even if the fixed weights of Eq.~\ref{equ_reward} occasionally misalign with specific out-of-distribution degradations, the tournament selection ensures that the ultimate output $I^*$ is anchored in human-aligned perceptual fidelity rather than spurious metric maximization.

\subsection{Dual-System Synergy and Memory Evolution}
\label{sec:synergy_evolution}
When P-MCTS yields a successful trajectory $\tau^*$ ($R(I_H) \ge \eta$), its state fingerprint is evaluated. Novel trajectories are newly incorporated into $\mathcal{M}$, while familiar ones update existing statistical profiles $\mathcal{K}_i$ (i.e., $n$, $\mu$, and $R_{\max}$). To maintain efficiency, pruning is triggered when $\mathcal{M}$ exceeds its capacity $M_{max}$. We formally execute this by evicting obsolete entries with the lowest visit counts $n$ (Least Frequently Used policy) and sub-optimal trajectories whose peak reward $R_{\max}$ falls below the memory's population mean. As $\mathcal{M}$ expands, the agent encounters familiar states more frequently, allowing more samples to be handled by verified memory shortcuts rather than full P-MCTS. This dynamic progressively reduces repeated planning for recurring degradation patterns. Ultimately, SEAR transforms expensive search trajectories into reusable expertise, providing a quality-efficiency tradeoff by amortizing search costs through memory reuse in long-term deployment.


\section{Experiments}
\label{sec:experiments}

\subsection{Experimental Setups}

\textbf{Implementation Details.} Following \cite{zhuintelligent2025}, we utilize pre-trained X-Restormer \cite{chen2024comparative}, HAT \cite{chen2025hat}, FBCNN \cite{jiang2021towards}, SwinIR \cite{liang2021swinir}, DiffBIR \cite{lin2024diffbir}, MPRNet \cite{mehri2021mprnet}, DRBNet \cite{ruan2022learning}, DehazeFormer \cite{song2023vision}, MAXIM \cite{tu2022maxim}, RIDCP \cite{wu2023ridcp}, Restormer \cite{zamir2022restormer}, and traditional operators.
To facilitate image degradation perception, we utilize a fine-tuned DepictQA \cite{you2024depicting} as the unified MLLM evaluator to assess visual quality and distortion types. For macro-task planning, GPT-4o \cite{achiam2023gpt} serves as the core LLM engine to orchestrate high-level restoration strategies. Key parameters are $c=0.8, \eta=0.55, \zeta=0.7, \omega=9$, and distance penalty $\alpha=0.05$.  Experiments use NVIDIA V100 GPUs (details in Appendix A).

\noindent \textbf{Datasets.}
The synthetic benchmarks (Groups A, B, and C) \cite{zhuintelligent2025} consist of 1,440 images derived from MiO100 \cite{kong2024towards} featuring 16 distinct combinations of mixed distortions. To assess generalization, the real-world dataset aggregates 100 pairs from I-Haze \cite{ancuti2018haze}, NH-Haze \cite{ancuti2020nh}, DRealSR \cite{wei2020component}, RealSR \cite{cai2019toward}, T-OLED \cite{zhou2021image}, SIDD \cite{abdelhamed2018high}, and LHP-Rain \cite{guo2023sky}, which will be made publicly available. 

\noindent \textbf{Baselines.} We evaluate SEAR against six AiO models (AirNet \cite{li2022all}, PromptIR \cite{potlapalli2023promptir}, MiOIR \cite{kong2024towards}, DA-CLIP \cite{luo2024controlling}, InstructIR \cite{conde2024instructir}, AutoDIR \cite{jiang2024autodir}) and two agentic frameworks (AgenticIR \cite{zhuintelligent2025}, 4KAgent \cite{zuo4kagent2025}). To ensure strategic parity and reproducibility, we unify toolsets and re-evaluate agentic baselines using the publicly released DepictQA module from AgenticIR \cite{zhuintelligent2025}, serving as a standardized proxy for their unreleased internal variant.

\noindent \textbf{Metrics.} 
Reference-based evaluation utilizes PSNR and SSIM \cite{wang2004image} computed on the luminance (Y) channel following the MiO100 protocol \cite{kong2024towards}. To ensure consistent comparison, all baseline methods are re-evaluated under this unified Y-channel setting. Perceptual similarity is assessed via LPIPS \cite{zhang2018unreasonable}. To evaluate restoration quality without ground truth, we report no-reference scores including MANIQA \cite{yang2022maniqa}, CLIP-IQA \cite{wang2023exploring}, and MUSIQ \cite{ke2021musiq} to quantify perceptual.

\begin{table}[t]
\caption{Quantitative comparison of multiple-degradation image restoration tasks on three subsets (Group A, B, and C) from the MiO100 dataset. The best and second best results are highlighted in \textbf{bold} and \underline{underline}, respectively.}
\label{table:quantitative_comparison_group}
\tabcolsep=0.1cm
\resizebox{0.97\textwidth}{!}{
\begin{tabular}{llcccccc}
\toprule
Datasets & Method & PSNR$\uparrow$ & SSIM$\uparrow$ & LPIPS$\downarrow$ & MANIQA$\uparrow$ & CLIP-IQA$\uparrow$ & MUSIQ$\uparrow$  \\
\midrule
\multirow{9}{*}{Group A}
& AirNet \cite{li2022all} & 20.8902 & 0.6675 & 0.4293 & 0.2650 & 0.3938 & 42.5047 \\
& PromptIR \cite{potlapalli2023promptir} & 21.4075 & 0.6701 & 0.4254 & 0.2693 & 0.3957 & 42.8456 \\
& MiOIR \cite{kong2024towards} & 21.3563 & 0.6834 & 0.4074 & 0.2626 & 0.4120 & 43.8445 \\
& DA-CLIP \cite{luo2024controlling} & 20.7522 & 0.6512 & 0.4225 & 0.2703 & 0.4155 & 44.6804 \\
& InstructIR \cite{conde2024instructir} & 20.2364 & 0.6409 & 0.4626 & 0.2630 & 0.3781 & 42.9907 \\
& AutoDIR \cite{jiang2024autodir} & 20.4909 & 0.6543 & 0.4138 & 0.2872 & 0.4172 & 48.9844 \\
& AgenticIR \cite{zhuintelligent2025} & 21.5321 & \textbf{0.6972} & 0.3110 & 0.3303 & 0.4842 & 59.8602 \\
& 4KAgent \cite{zuo4kagent2025} & \underline{21.6468} & 0.6936 & \underline{0.3082} & \underline{0.3403} & \underline{0.5017} & \underline{61.1519} \\
& SEAR & \textbf{21.8042} & \underline{0.6961} & \textbf{0.3019} & \textbf{0.3411} & \textbf{0.5039} & \textbf{61.2882}\\
\midrule
\multirow{9}{*}{Group B}
& AirNet \cite{li2022all} & 20.1531 & 0.7015 & 0.3753 & 0.2946 & 0.4184 & 47.4868 \\
& PromptIR \cite{potlapalli2023promptir} & 20.3621 & 0.7146 & 0.3550 & 0.2961 & 0.4208 & 47.5886 \\
& MiOIR \cite{kong2024towards} & 20.6956 & \underline{0.7244} & 0.3477 & 0.2830 & 0.4325 & 48.5851 \\
& DA-CLIP \cite{luo2024controlling} & 20.1534 & 0.6741 & 0.3968 & 0.2873 & 0.4111 & 48.4284 \\
& InstructIR \cite{conde2024instructir} & 20.8317 & 0.6761 & 0.4133 & 0.3016 & 0.4006 & 48.9953 \\
& AutoDIR \cite{jiang2024autodir} & 20.9553 & 0.6897 & 0.3837 & 0.2925 & 0.4204 & 49.0215 \\
& AgenticIR \cite{zhuintelligent2025} & \underline{21.7160} & 0.7236 & 0.3025 & 0.3311 & 0.4871 & 59.5741 \\
& 4KAgent \cite{zuo4kagent2025} & 21.5416 & 0.7175 & \underline{0.2958} & \underline{0.3434} & \textbf{0.5226} & \textbf{61.4795} \\
& SEAR & \textbf{22.1332} & \textbf{0.7251} & \textbf{0.2890} & \textbf{0.3447} & \underline{0.5134} & \underline{60.6520} \\
\midrule
\multirow{9}{*}{Group C}
& AirNet \cite{li2022all} & 20.5462 & 0.5983 & 0.5618 & 0.2026 & 0.3221 & 29.9750 \\
& PromptIR \cite{potlapalli2023promptir} & \underline{20.6046} & \underline{0.6012} & 0.5584 & 0.2059 & 0.3189 & 29.8850  \\
& MiOIR \cite{kong2024towards} & 20.5308 & \textbf{0.6015} & 0.5725 & 0.1962 & 0.3171 & 30.1058 \\
& DA-CLIP \cite{luo2024controlling} & 19.5827 & 0.5683 & 0.5948 & 0.1921 & 0.3330 & 29.5910 \\
& InstructIR \cite{conde2024instructir} & 19.4498 & 0.5781 & 0.5756 & 0.1890 & 0.3103 & 29.2901 \\
& AutoDIR \cite{jiang2024autodir} & 19.1069 & 0.5716 & 0.5677 & 0.2113 & 0.3226 & 35.7965 \\
& AgenticIR \cite{zhuintelligent2025} & 20.1974 & 0.5875 & 0.4341 & 0.2719 & 0.4325 & 51.7533 \\
& 4KAgent \cite{zuo4kagent2025} & 20.3078 & 0.5798 & \underline{0.4223} & \textbf{0.2929} & \textbf{0.4737} & \underline{52.6504} \\
& SEAR & \textbf{20.6121} & 0.5927 & \textbf{0.4173} & \underline{0.2742} & \underline{0.4420} & \textbf{52.7338} \\
\bottomrule
\end{tabular}}
\end{table}

\subsection{Comparison with State-of-the-Arts}

\begin{figure}[ht]
  \centering
  \includegraphics[width=\textwidth]{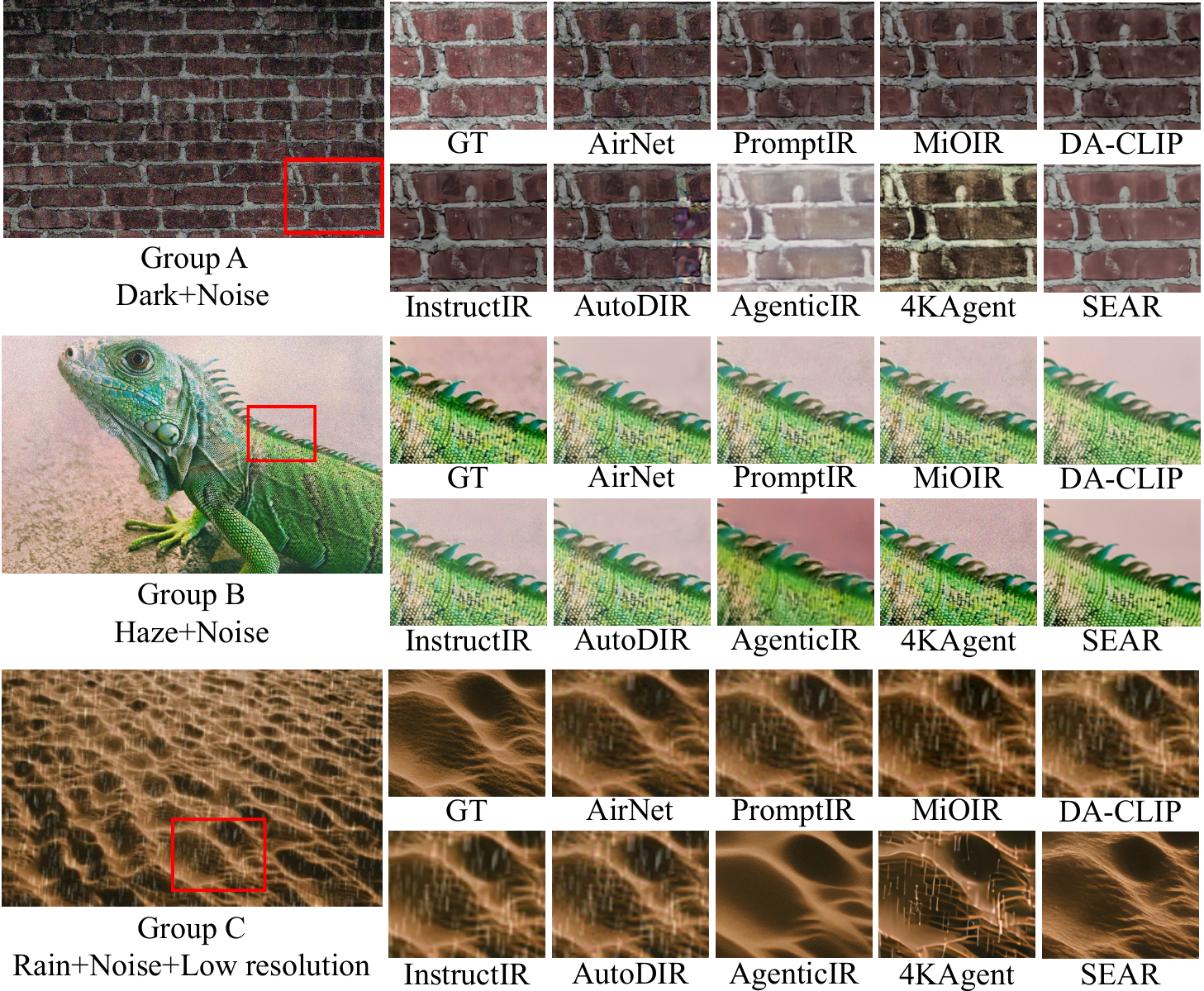}
  \caption{Qualitative comparison on synthetic datasets \cite{zhuintelligent2025}.}
  \label{FIG_visual_result_syn}
\end{figure}

\textbf{Quantitative Comparison.}
Table~\ref{table:quantitative_comparison_group} shows that SEAR consistently outperforms AiO and agentic baselines across nearly all benchmarks. Notably, it achieves PSNR gains of 0.42 dB and 0.30 dB in Groups B and C, where out-of-domain patterns elude memory-less agents \cite{zhuintelligent2025,zuo4kagent2025}. SEAR also leads in perceptual metrics (LPIPS and MUSIQ), underscoring high-fidelity reconstruction in unseen scenarios. Such consistency confirms that SEAR's self-evolving architecture bridges the gap between fixed heuristics and dynamic restoration requirements.

\noindent \textbf{Qualitative Comparisons.}
Fig.~\ref{FIG_visual_result_syn} presents qualitative comparisons on the synthetic MiO100 dataset. AiO models struggle in mixed degradation scenarios, frequently failing to reconcile conflicting distortions and introducing visible artifacts. Although AgenticIR produces comparatively cleaner outputs, its greedy search strategy often results in over-smoothing, leading to the loss of characteristic structures, such as the sand ripples in Group~C. 4KAgent further suffers from noticeable residual noise and perceptual distortions. In contrast, SEAR consistently preserves fine-grained details across all categories, accurately recovering brick mortar, lizard scales, and intricate dune textures, while effectively suppressing noise and removing rain. Additional results are in Appendix B.

\begin{wraptable}{r}{0.4\linewidth}
\vspace{-15pt}
\centering
\caption{Efficiency analysis on the Group~B dataset \cite{zhuintelligent2025}.}
\label{tab:comparison_groupb}
\renewcommand{\arraystretch}{1.1}
\resizebox{\linewidth}{!}{
\begin{tabular}{l|ccc}
\toprule
Method & PSNR & Time & Tool Calls \\
\midrule
AgenticIR & 21.72 & 1.09 & 6.11 \\
4KAgent   & 21.54 & 2.55 & 8.26 \\
SEAR & 22.13 & 1.98 & 8.15 \\
\bottomrule
\end{tabular}
}
\end{wraptable}

\noindent \textbf{Efficiency Analysis.} 
Table~\ref{tab:comparison_groupb} evaluates efficiency using average inference time (min) and tool calls per image. SEAR balances fidelity and efficiency by amortizing planning costs through episodic memory. Compared to AgenticIR, SEAR improves PSNR by 0.42 dB with limited runtime and tool overhead. SEAR also outperforms 4KAgent in restoration quality and efficiency, reducing execution time by 22.4\% with comparable tool usage.


\begin{table}[t!]
\centering
\caption{Quantitative comparison on the real-world dataset.}
\label{table:quantitative_comparison_real}
\tabcolsep=0.1cm
\resizebox{0.9\textwidth}{!}{
  \begin{tabular}{lcccccc}
  \toprule
  Method & PSNR$\uparrow$ & SSIM$\uparrow$ & LPIPS$\downarrow$ & MANIQA$\uparrow$ & CLIP-IQA$\uparrow$ & MUSIQ$\uparrow$ \\
  \midrule
  AirNet \cite{li2022all} & 23.3067 & 0.7471 & 0.4484 & 0.2356 & 0.3272 & 35.2690 \\
  PromptIR \cite{potlapalli2023promptir} & 23.8647 & \underline{0.7542} & 0.4341 & 0.2405 & 0.3270 & 35.4811 \\
  MiOIR \cite{kong2024towards} & 24.0043 & 0.7472 & 0.4239 & 0.2394 & 0.3507 & 36.8458 \\
  DA-CLIP \cite{luo2024controlling} & 23.7313 & 0.7437 & 0.4273 & 0.2514 & 0.3330 & 36.4113 \\
  InstructIR \cite{conde2024instructir} & \textbf{26.1477} & \textbf{0.7639} & 0.4237 & 0.2556 & 0.3412 & 37.8236 \\
  AutoDIR \cite{jiang2024autodir} & 20.7470 & 0.6640 & 0.4792 & 0.2466 & 0.3148 & 42.6789 \\
  AgenticIR \cite{zhuintelligent2025} & 23.9280 & 0.7214 & 0.3730 & 0.3151 & 0.4512 & 52.9103 \\
  4KAgent \cite{zuo4kagent2025} & 23.6295 & 0.7242 & \underline{0.3569} & \textbf{0.3200} & \underline{0.4513} & \underline{54.4647} \\
  SEAR & \underline{24.4078} & 0.7425 & \textbf{0.3371} & \underline{0.3174} & \textbf{0.4519} & \textbf{54.6686} \\
  \bottomrule
  \end{tabular}}
\end{table}

\begin{figure}[t!]
  \centering
  \includegraphics[width=\textwidth]{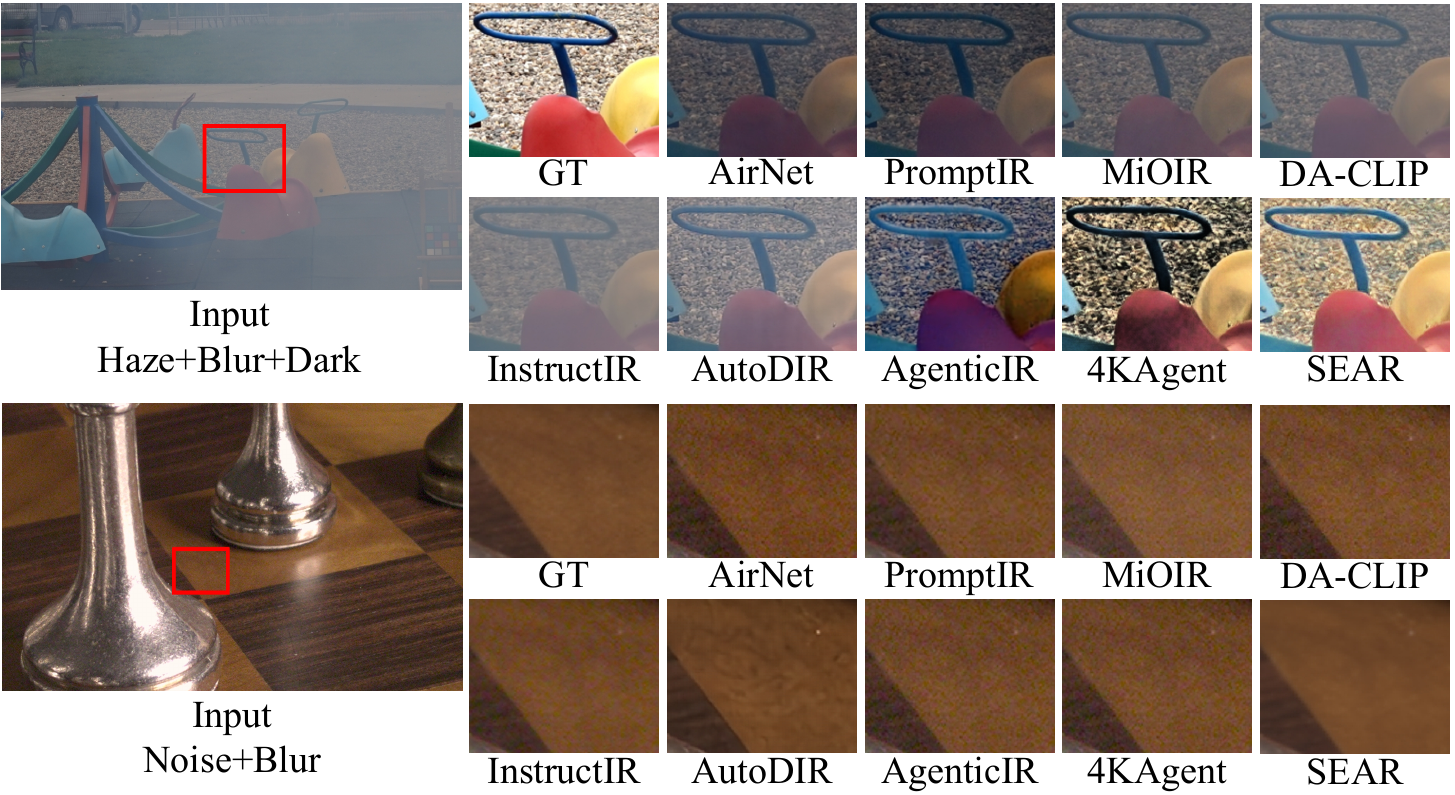}
  \caption{Qualitative comparison on the real-world dataset.}
  \label{FIG_visual_result_real}
\end{figure}

\subsection{Real-World Generalization} 
Evaluation on the real-world dataset demonstrates the robustness of SEAR in real-world scenarios. As shown in Table \ref{table:quantitative_comparison_real}, SEAR attains leading performance across perceptual metrics, including LPIPS and CLIP-IQA. Despite superior PSNR, pixel-fidelity models like InstructIR frequently produce oversmoothed outputs with limited perceptual detail.
Visual results in Fig. \ref{FIG_visual_result_real} confirm these observations. Particularly in ``Haze+Blur'' and ``Noise+Blur'' scenarios, SEAR restores natural textures and effectively suppresses artifacts, whereas baselines exhibit residual distortions or severe color casts. By optimizing restoration trajectories, SEAR handles complex real-world degradations with optimal perceptual quality, solving mixed patterns that challenge static models.

\begin{table}[t]
\centering
\caption{Ablation study on the Group B dataset. ``Tool Calls'' denotes the average number of restoration tool invocations per image, reflecting inference efficiency.}
\label{tab:ablation_main}
\resizebox{\linewidth}{!}{
\begin{tabular}{l|cccccc|c}
\toprule
Configuration & PSNR$\uparrow$ & SSIM$\uparrow$ & LPIPS$\downarrow$ & MANIQA$\uparrow$ & CLIP-IQA$\uparrow$ & MUSIQ$\uparrow$  & Tool Calls$\downarrow$  \\ \midrule
a. SEAR (Full) & \textbf{22.1332} & 0.7251 & \textbf{0.2890} & 0.3447 & 0.5134 & 60.6520 & 8.15 \\
b. w/o Memory & 22.0222 & \textbf{0.7261} & 0.2903 & 0.3492 & 0.5124 & 61.0802 & 16.75 \\
c. w/o P-MCTS & 20.8395 & 0.7156 & 0.3046 & 0.3335 & 0.5068 & 60.3226 & \textbf{4.08} \\
d. w/o Macro-Task & 19.5378 & 0.6367 & 0.3728 & \textbf{0.3926} & \textbf{0.5488} & \textbf{64.4798} & 71.76\\
e. Vanilla MCTS & 22.1252 & 0.7233 & 0.2893 & 0.3480 & 0.5128 & 60.8131 & 9.93\\
f. w/o Fingerprint & 21.0774 & 0.6848 & 0.3335 & 0.3458 & 0.4885 & 59.9639 & 11.49
\\ \bottomrule
\end{tabular}
}
\end{table}

\begin{figure}[t!]
  \centering
  \begin{minipage}[t]{0.44\textwidth}
    \centering
    \includegraphics[width=\linewidth]{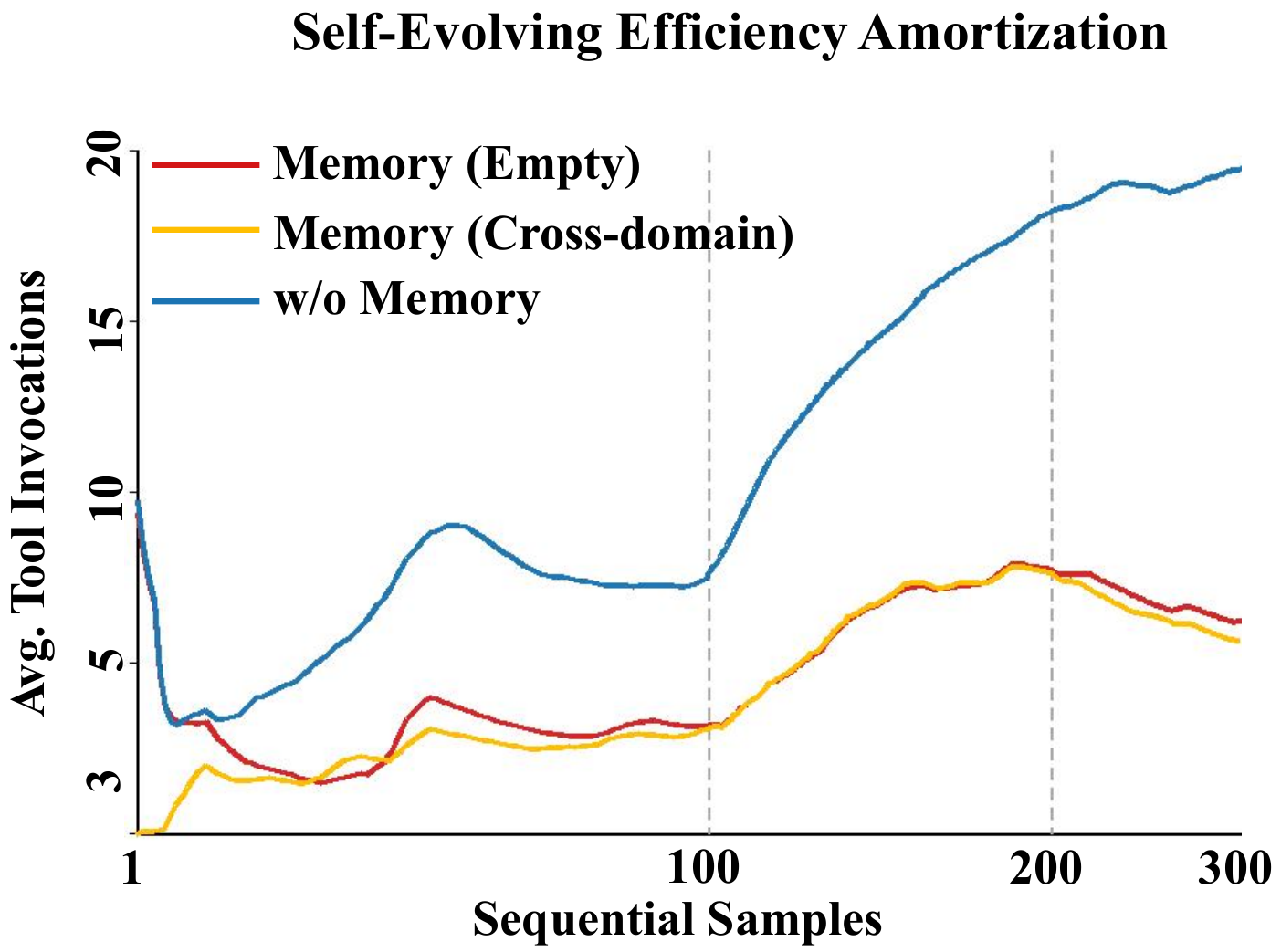}
    \caption{Dynamic self-evolution.}
    \label{FIG_visual_memory}
  \end{minipage}
  \hfill
  \begin{minipage}[t]{0.54\textwidth}
    \centering
    \includegraphics[width=\linewidth]{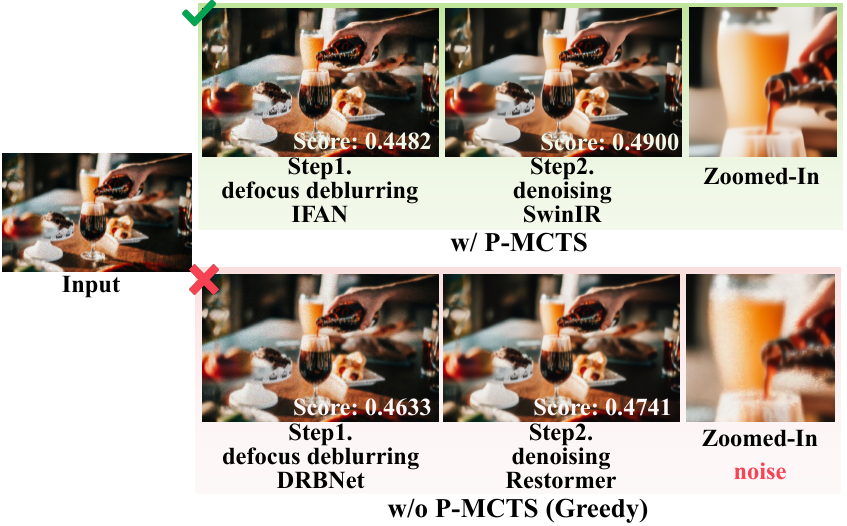}
    \caption{Ablation study of P-MCTS.}
    \label{FIG_visual_greedy}
  \end{minipage}
\end{figure}

\begin{figure}[t]
  \centering
  \includegraphics[width=\textwidth]{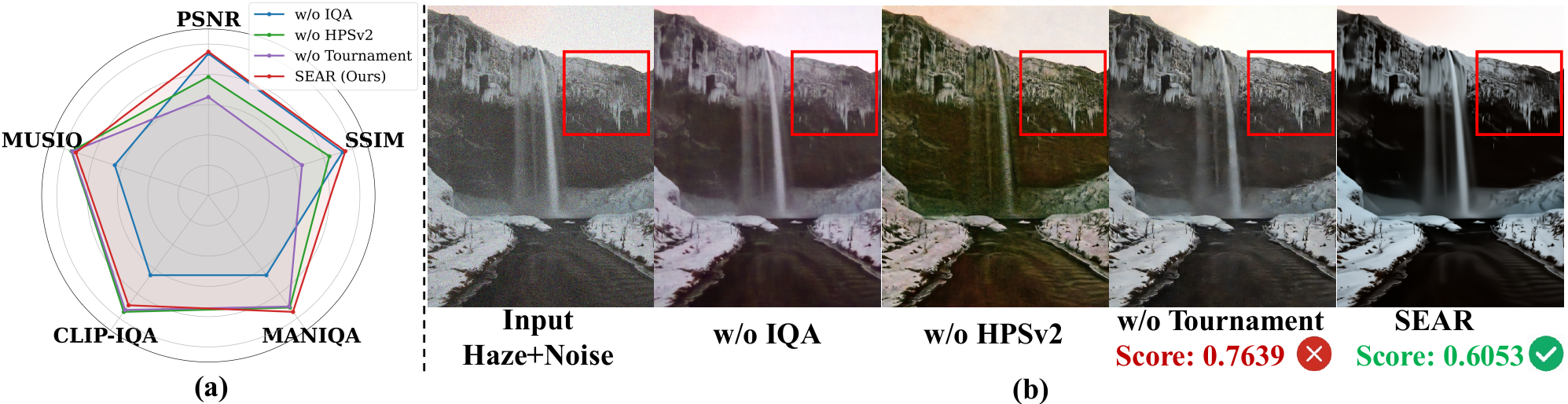}
  \caption{Reward design ablation. (a) Radar comparison of reward components. (b) Visual analysis: tournament selection prioritizes perceptual quality over inflated scores.}
\label{FIG_visual_metric_exploitation}
\end{figure}

\subsection{Ablation Study}
\label{subsection:ablation_study}

\noindent \textbf{Synergy of Dual-Process Systems.} Table \ref{tab:ablation_main} (Configs. b, c) confirms that SEAR's dual-process architecture is vital for balancing fidelity and efficiency. Removing the memory bank forces the agent to search without dynamic memory guidance, doubling tool calls from 8.15 to 16.75, demonstrating that memory-guided inference is essential for efficient execution. 
Fig.~\ref{FIG_visual_memory} illustrates the dynamic self-evolution of SEAR. 
While the cold-start SEAR (red) initially behaves like the deliberative search baseline (blue), it quickly distills planning into efficient execution. Cross-domain initialization (yellow) further accelerates convergence, demonstrating the transferability of distilled trajectories, while the eventual convergence of both settings highlights the robustness of the self-evolving architecture.
Conversely, replacing P-MCTS with greedy heuristics reduces tool calls to 4.08 but causes a 1.29 dB drop in PSNR. As shown in Fig. \ref{FIG_visual_greedy}, greedy heuristics like premature sharpening trigger irreversible artifact accumulation. 

\noindent \textbf{Memory Routing Analysis.}
To further verify that SEAR does not blindly reuse retrieved trajectories, we report memory routing statistics under cold-start deployment in Table~\ref{tab:memory_routing}. For each input, the Intuitive Executor first retrieves candidate trajectories according to degradation-aware fingerprints. The retrieved trajectory is executed only when it passes the reliability gate; otherwise, the sample is routed to the Deliberate Planner for P-MCTS search. As shown in Table~\ref{tab:memory_routing}, SEAR achieves high retrieval rates on both synthetic and real-world benchmarks, while a substantial portion of uncertain cases is still routed to System~2. This execute-then-gate mechanism confirms that the memory bank serves as a verified shortcut rather than an unconditional cache.

\begin{table}[t]
\centering
\caption{Memory routing statistics under cold-start deployment.}
\label{tab:memory_routing}
\setlength{\tabcolsep}{6pt}
\footnotesize
\begin{tabular}{lccc}
\toprule
Dataset & Memory Hit & Accepted Shortcut & Routed to System~2 \\
\midrule
Group A & 97\% & 67\% & 33\%  \\
Group B & 98\% & 62\% & 38\%   \\
Group C & 94\% & 43\% & 57\%  \\
Real-world & 83\% & 41\% & 59\%   \\
\bottomrule
\end{tabular}
\end{table}

\noindent \textbf{Efficiency of Hierarchical Planning.} Table \ref{tab:ablation_main} (Configs. d and e) validates the necessity of hierarchical planning. Removing the macro-task forces a flat search across the entire tool set, increasing tool calls to 71.76 while degrading PSNR to 19.54 dB, clearly showing the efficiency and effectiveness benefits of hierarchical constraints. Furthermore, reverting to vanilla MCTS marginally degrades PSNR while inflating tool invocations from 8.15 to 9.93, demonstrating that MLLM-guided action pruning effectively curtails redundant exploration.

\begin{table}[t]
\centering
\caption{Hyperparameter sensitivity analysis on the real-world dataset.}
\label{tab:hyper_sensitivity}
\setlength{\tabcolsep}{4pt} 
\footnotesize
\resizebox{0.85\linewidth}{!}{
\begin{tabular}{lcc}
\toprule
$c$ & PSNR & SSIM \\ \midrule
0.4 & 24.16 & 0.7373 \\
0.6 & 24.08 & 0.7257 \\
0.8 & \textbf{24.41} & \textbf{0.7425} \\
1.0 & 24.26 & 0.7396 \\
1.2 & 24.14 & 0.7376 \\ \bottomrule
\end{tabular}
\hspace{12pt} 
\begin{tabular}{lcc}
\toprule
 & PSNR & SSIM \\ \midrule
$\zeta=0.6$ & 23.89 & 0.7320 \\
\textbf{$\zeta=0.7$} & \textbf{24.41} & \textbf{0.7425} \\
$\zeta=0.8$ & 24.20 & 0.7424 \\ \midrule
$\eta=0.45$ & 24.05 & 0.7270 \\
\textbf{$\eta=0.55$} & \textbf{24.41} & \textbf{0.7425} \\
$\eta=0.65$ & 24.20 & 0.7389 \\ \bottomrule
\end{tabular}
\hspace{12pt}
\begin{tabular}{lcc}
\toprule
$\omega$ & PSNR & SSIM \\ \midrule
7 & 24.18 & 0.7349 \\
8 & 24.39 & \textbf{0.7540} \\
9 & \textbf{24.41} & 0.7425 \\
10 & 24.31 & 0.7456 \\
11 & 24.16 & 0.7318 \\
12 & 24.28 & 0.7360 \\ \bottomrule
\end{tabular}
}
\end{table}

\noindent \textbf{Importance of State Fingerprint.} Table \ref{tab:ablation_main} (Config. f) shows that replacing the semantic fingerprint with IQA-only descriptors reduces PSNR by 1.05 dB, indicating that simple quality metrics cannot capture heterogeneous degradations. Therefore, MLLM-based state abstraction is essential for accurate retrieval.

\noindent \textbf{Perceptual Alignment Analysis.} Fig.~\ref{FIG_visual_metric_exploitation} shows that ablating terminal hybrid reward $R(I_H)$ components evaluates the trade-off between pixel accuracy and perceptual quality. Using the IQA-only reward preserves unsupervised metrics (e.g., NIQE, MUSIQ) but ignores aesthetic fidelity, whereas relying solely on HPSv2 favors human-aligned aesthetics at the cost of structural accuracy. Removing either component consistently degrades both quantitative and qualitative performance, confirming that the combined hybrid reward best guides restoration. To prevent trajectories that achieve high reward scores but produce visually unsatisfactory results, we apply the MLLM Tournament as a final perceptual filter. This ensures that SEAR prioritizes both metric consistency and human-aligned visual quality, achieving robust coverage across all dimensions.

\noindent \textbf{Hyperparameter Sensitivity.} Table \ref{tab:hyper_sensitivity} confirms SEAR's structural robustness across core parameters. Specifically, the exploration constant $c=0.8$ provides a favorable balance between search diversity and convergence, while the distillation threshold $\eta=0.55$ effectively gates memory quality against the rate of expertise accumulation. Furthermore, the shortcut threshold $\zeta=0.7$ supports reliable shortcut execution by bypassing deliberation for recurring patterns without performance degradation. $\omega=9$ represents the optimal search budget multiplier in P-MCTS, as it effectively saturates restoration performance without redundant computation. Detailed tuning strategies are provided in Appendix A.






\section{Conclusion}
\label{sec:conclusion}
To address the greedy planning and episodic amnesia in current restoration agents, we proposed SEAR, a dual-process framework that synergizes deliberate planning with intuitive execution. By coupling P-MCTS with self-evolving episodic memory, SEAR enables long-horizon trajectory optimization while distilling expensive search experiences into adaptive, reusable expertise. This architecture allows the system to progressively shift from resource-intensive deliberation to efficient, memory-driven inference as experience accumulates. Experiments on synthetic and real-world benchmarks demonstrate that SEAR achieves competitive restoration quality, pointing to a practical pathway toward self-evolving agentic restoration systems in complex real-world environments.

\paragraph{\textbf{Limitations.}}
Although SEAR achieves strong restoration performance, it still has two limitations. First, SEAR mainly handles unseen degradation combinations and severities covered by a predefined restoration tool library, and its memory retrieval depends on degradation diagnosis for state fingerprint construction. Inaccurate diagnosis or missing restoration tools may affect planning under severe or out-of-distribution degradations. Second, the initial P-MCTS stage introduces cold-start overhead before sufficient high-quality trajectories are accumulated in memory. Future work will explore broader tool libraries, more discriminative fingerprints, and stronger perception-aligned evaluation for more diverse restoration scenarios.


\section*{Acknowledgements}
This work is supported by the National Natural Science Foundation of China under Grant No. 62406313, the Postdoctoral Fellowship Program of the China Postdoctoral Science Foundation under Grant No. YJB20250283, and the National Key Laboratory of Space Integrated Information System under Grant No. E5GF5910.

%
%
\bibliographystyle{splncs04}
\bibliography{main}

\end{document}